\documentclass[runningheads]{llncs}
\usepackage{graphicx}
\usepackage{dsfont}
\usepackage{amsmath}
\usepackage{float}
\usepackage{mathtools}
\usepackage[T1]{fontenc}
\usepackage{amsmath}
\usepackage{caption}
\usepackage{subcaption}
\usepackage{textalpha}
\usepackage{dsfont}
\usepackage{amsmath}
\usepackage{float}
\usepackage{multirow}
\usepackage{booktabs} % For formal tables
\usepackage{hhline}
\usepackage{longtable}
\usepackage{caption}
\usepackage{xcolor,pifont}
\usepackage{multicol}
\usepackage{tabularx}
\usepackage{amssymb}% http://ctan.org/pkg/amssymb
\usepackage{pifont}% http://ctan.org/pkg/pifont
\usepackage{adjustbox}
\usepackage{lipsum}
\usepackage{tikz}
\usepackage{enumitem}
\usepackage{booktabs}
\usepackage{xcolor,pifont}
\usepackage{amssymb}% http://ctan.org/pkg/amssymb
\usepackage{pifont}% http://ctan.org/pkg/pifont
\newcommand*\colourcheck[1]{%
  \expandafter\newcommand\csname #1check\endcsname{\textcolor{#1}{\ding{52}}}%
}

\usepackage[colorlinks=true, citecolor=citecolor, linkcolor=linkcolor]{hyperref}
\definecolor{citecolor}{HTML}{0071BC}
\definecolor{linkcolor}{HTML}{ED1C24}
\usepackage{verbatim}
\usepackage[capitalize]{cleveref}

\begin{document}
\def\thefootnote{*}\footnotetext{These authors contributed equally to this work}
%%Paper ID: 53
\title{UnSupDLA: Towards Unsupervised Document Layout Analysis}
%\author{Tahira Shehzadi*\inst{1,2,3}\orcidID{0000-0002-7052-979X} \and
%Didier Stricker\inst{1,2,3} \and
%Muhammad Zeshan Afzal\inst{1,2,3}\orcidID{0000-0002-0536-6867}}
%
%\authorrunning{T. Shehzadi et al.}
%
%\institute{Department of Computer Science, Technical University of Kaiserslautern, 67663 Kaiserslautern, Germany \and
%Mindgarage, Technical University of Kaiserslautern, 67663 Kaiserslautern, Germany \and
%German Research Institute for Artificial Intelligence (DFKI), 67663 Kaiserslautern, Germany\\
%\email{\{tahira.shehzadi@dfki.de\}
%\email{\{firstname_middlename.lastname@dfki.de\}}}
%
%\maketitle              % typeset the header of the contribution
%
%\author{Talha Uddin Sheikh*\inst{1,2,3}\orcidID{0009-0004-9156-5679} \and
%Tahira Shehzadi*\inst{1,2,3}\orcidID{0000-0002-7052-979X} \and
%Khurram Azeem Hashmi\inst{1,2,3}\orcidID{0000-0003-0456-6493} \and
%Didier Stricker\inst{1,2,3} \and
%Muhammad Zeshan Afzal\inst{1,2,3}\orcidID{0000-0002-0536-6867}}
\author{Talha Uddin Sheikh*\inst{1,2,3}\orcidID{0009-0004-9156-5679} \and
Tahira Shehzadi*\inst{1,2,3}\orcidID{0000-0002-7052-979X} \and
Khurram Azeem Hashmi\inst{1,2,3}\orcidID{0000-0003-0456-6493} \and
Didier Stricker\inst{1,2,3} \and
Muhammad Zeshan Afzal\inst{1,2,3}\orcidID{0000-0002-0536-6867}}
\authorrunning{T. Sheikh et al.}
\institute{Department of Computer Science, Technical University of Kaiserslautern, Germany \and
Mindgarage, Technical University of Kaiserslautern, Germany \and
 German Research Institute for Artificial Intelligence (DFKI), 67663 Kaiserslautern, Germany\\
\email{\{firstname\_middlename.lastname@dfki.de\}}}
%\def\thefootnote{*}\footnotetext{These authors contributed equally to this work}
%\newpage
%
\maketitle              % typeset the header of the contribution

\begin{abstract}
Document layout analysis is a key area in document research, involving techniques like text mining and visual analysis. Despite various methods developed to tackle layout analysis, a critical but frequently overlooked problem is the scarcity of labeled data needed for analyses. With the rise of internet use, an overwhelming number of documents are now available online, making the process of accurately labeling them for research purposes increasingly challenging and labor-intensive. Moreover, the diversity of documents online presents a unique set of challenges in maintaining the quality and consistency of these labels, further complicating document layout analysis in the digital era. To address this, we employ a vision-based approach for analyzing document layouts designed to train a network without labels. Instead, we focus on pre-training, initially generating simple object masks from the unlabeled document images. These masks are then used to train a detector, enhancing object detection and segmentation performance. The model's effectiveness is further amplified through several unsupervised training iterations, continuously refining its performance. This approach significantly advances document layout analysis, particularly precision and efficiency, without labels.
\end{abstract}
\section{Introduction}
Document layout analysis (DLA) has always been a key challenge in computer vision and document understanding. Historically, the field has developed diverse methodologies~\cite{DLA_survey56}, ranging from traditional classical techniques~\cite{VoronoiAD_09,ANN_survey_TPAMI_05,tab6f5} to more contemporary, learning-based models~\cite{layoutMV3,LayoutParser9}. The advancement of technologies such as convolutional neural networks (CNNs) has marked a notable improvement in the precision and functionality of these models, showing a significant evolution in the approach to DLA~\cite{cdec,docsegtr,huang2022layoutlmv3,bridging_per3,shehzadi2024hybrid6,shehzadi2024endtoend7,ehsan_semi8}. As technology advances, there has been a corresponding change in the complexity of documents, particularly in the digital domain. This shift is most evident in business environments, where documents come in increasingly varied and complex formats~\cite{DLA_survey56,survey_dnn5}. These developments present a new set of challenges, requiring models that are accurate and adaptable enough to adjust to a wide range of document types and layouts. In response to this dynamic landscape, the strategies employed in DLA have been continuously refined and improved. The focus has expanded to include the accuracy of analysis and the adaptability to handle the diverse array of modern document formats. This ongoing advancement in DLA methods underscores the importance and persistent relevance of the field in the broader context of document understanding and computer vision research. As documents continue to evolve, so will the techniques and technologies in DLA, ensuring that it remains an essential and ever-progressing study area~\cite{DAD_IJDAR22}.

Previously, classical rule-based methods were employed for document layout analysis~\cite{RecogTable5,TTsurvey8,extractTab9,DEA38,TSsurvey32}. More recently, it's been approached as a Document Object Detection (DOD) problem, employing vision-based object detection models~\cite{trainTD5,DEA38,TSsurvey32,shehzadi2023object5,Xi17,semimask4,He761,sparse_semi_detr2}. Researchers have also combined sequence and language models with object detection for better accuracy~\cite{layoutMV3}. However, there's an overlooked issue. Unconventional document formats require labor-intensive annotation for traditional supervised methods. So, unsupervised approaches have become important. Implementing unsupervision in DOD is challenging because images contain multiple document objects of different classes, and treating each image as a class isn't effective. However, it's worth noting that these salient object detection methods~\cite{tokencut_TPAMI23} are specifically designed to locate a single object, typically the most prominent one, and may not be suitable for handling real-world document images containing multiple objects and complex layouts. This raises questions about the effectiveness of unsupervision in document segmentation.

In this paper, we identify and localize graphical elements within documents without labels. In the initial phase of unsupervised training, We use unlabeled data, which lacks specific information about the locations and types of objects in the documents. We generate initial layout masks based on features from a self-supervised DINO~\cite{DINO_selfsup3}. We analyze patch-wise similarities for images with multiple objects and use Normalized Cuts (NCut) to isolate a mask for each object, repeating this multiple times for multiple objects. Later, we apply a loss drop strategy in the detector training to improve performance. The model undergoes several iterations of unsupervised training for further refinement. Previous research has shown self-supervised vision-based methods~\cite{DINO_selfsup3,SelfDoc_CVPR21} to be less effective for DLA tasks because they require direction from learned text and layout embeddings. Yet, we propose that unsupervised learning employs visual representation. The visual features generate masks that provide a preliminary idea of where objects might be located within the documents, serving as a starting point for further analysis.
In short, our approach does not rely on layout information from pre-trained text recognition models. Instead, we use the inherent visual information within documents as a layout guide for learning visual representations.

In summary, the contributions of our paper are as follows:
\begin{itemize}
\item[$\bullet$] A vision-based unsupervised learning framework aims to train the detector to perform document layout analysis. This approach recognizes and analyzes the layout of documents autonomously.
\item[$\bullet$] A layout-guided strategy that generates initial layout masks using visual features for document segmentation.
\item[$\bullet$] An efficient unsupervised learning approach that learns about different document objects to minimize data use. It can be used as a pre-training model for document analysis.
\end{itemize}
We organize the content of the paper as follows. We begin with a thorough review of existing literature in Section~\ref{sec:RW}. Then, in Section~\ref{sec:method}, we detail the methodology. Section~\ref{sec:exp} is dedicated to the discussion of our experiments and the results obtained. In Section~\ref{sec:AS}, we conduct an ablation analysis. Finally, we conclude our paper in Section~\ref{sec:conclusion} with our final thoughts and findings.

\section{Related Work}
\label{sec:RW}
\subsection{Fully-Supervised Document Understanding}
Recent advancements in deep learning methodologies have broadened their applications, extending from healthcare~\cite{shehzadi_IEEE_I9,Protein10}, traffic analysis~\cite{wajahatCC8}, to document analysis~\cite{continuaLR45,Real_DICls4,rethink78,naik86,cas10,TSRkhurm4}.
In recent years, the idea of Document Understanding (DU) has expanded to include many different challenges and tasks related to Document Intelligence systems~\cite{DUEED21}. This includes, but is not limited to, Key Information Extraction~\cite{funsd_ICDARW19,CORDAC19,Kleister_ICDAR21}, Document Classification~\cite{EvaluationOD_15}, Document Layout Analysis~\cite{PubLayNet3,HJ_dataset_CVPRW20}, Question Answering~\cite{docvqa_WACV21,hierarchical_PR23}, and Machine Reading Comprehension~\cite{visualmrc_AAAI21}, particularly when dealing with Visually Rich Documents (VRDs) as opposed to simple text or basic image-text combinations. Leading DU systems predominantly utilize extensive pre-training to merge visual and textual elements~\cite{docformer_ICCV21,layoutMV3,SelfDoc_CVPR21,unidoc21,Doc2Graph_ECCV22}. However, methods like Donut~\cite{ocr_free_ECCV22} and Dessurt~\cite{Dessurt_ECCV22} focus more on enhancing visual features using synthetic generation techniques~\cite{docsynth_ICDAR21,synthtiger_ICDAR21,ContentAS_TPAMI21} for effective layout representation during document pre-training. 

\subsection{Fully-Supervised Document Layout Analysis}
DLA has emerged as a key application in data utilization, focusing on optimizing storage and handling of vast amounts of information~\cite{DLA_survey56}. The field has transformed with the introduction of deep learning and Convolutional Neural Networks (CNN), leading to a shift in document layout segmentation~\cite{DeepDeSRT3,LayoutParser9,hybridvbert,Da_2023_ICCV,SwinDoc} towards a Document Object Detection. The development of extensive DLA benchmarks~\cite{PubLayNet3,HJ_dataset_CVPRW20} has made it easier for deep learning techniques to be applied in this field. Biswas et.al~\cite{Iseg_Biswas_IJDAR21} has considered DLA as an instance-level segmentation task that is crucial for identifying bounding boxes and segmentation masks in pages with overlapping elements. Transformer-based methods~\cite{layoutMV3,shehzadi_semi-detr_table1} have recently achieved improved results in DLA, particularly for large-scale document datasets, though they still face challenges in smaller datasets. Innovative language-based methods like LayoutLMv3~\cite{layoutMV3} and UDoc~\cite{unidoc21} have shown impressive results on the PubLayNet benchmark but struggle with more complex layouts and smaller data samples. 
\subsection{Advancements in Self-Supervised Learning}
In the evolving field of computer vision, researchers have been concentrating on understanding complex visual details from different images. This led to the development of data-driven machine learning models, for extracting and correlating features, to meet increasingly complex demands. Advanced networks require a lot of data. This makes data annotation very important, leading to many self-supervised learning strategies. MoCo~\cite{momentum_unsup_CVPR20} introduced a novel approach in contrastive learning settings, utilizing exponential moving averages and large memory banks for weight updates. Building on this, SimCLR~\cite{contrastLearn_ICML20} proposed using larger batch sizes as an alternative to memory banks. DINO~\cite{DINO_selfsup3} brought the concept of self-supervision to vision transformers~\cite{Transformers_IR_56}. MoCov2~\cite{improved_cv2_20} and SwAV~\cite{unsupervised_CCA_20} subsequently achieved remarkable results within this self-supervised framework. Alternatively, BYOL~\cite{BYOL_NIPS20} and SimSiam~\cite{rep_learn_CVPR45} approached the problem by treating different sections of the same image as analogous pairs, moving away from traditional contrastive learning. Additionally, masked autoencoders~\cite{Masked_Autoencoders_21} have revitalized classic autoencoder techniques by incorporating a masking strategy for learning representations through reconstruction.

Despite the remarkable success of supervised object detection techniques such as Mask RCNN~\cite{mask-rcnn84}, Yolo~\cite{yolos6}, Retinanet~\cite{retinaNet68}, and DETR~\cite{detr34}, their self-supervised alternatives have been somewhat limited in scope until recently. Recent advancements have seen the development of end-to-end self-supervised object detection models like UP-DETR~\cite{updetr23} and DETReg~\cite{DETReg7}, as well as backbone pre-training strategies such as Self-EMD~\cite{Self_emd_20} and Odin~\cite{object_discovery_ECCV22}. %While significant research has been done on self-supervised learning, unsupervised methods still need to be explored. While some attempts have been made at self-supervised document segmentation~\cite{li2021selfdoc}, these methods have yet to utilize unsupervised visual representation or guidance effectively. This paper aims to fill this gap by introducing unsupervised document analysis. 
While significant research has been done on self-supervised learning, unsupervised methods still need to be explored. While some attempts have been made at unsupervised document analysis~\cite{icpr5,cross_domain7}, these methods have yet to improve effectively. This paper aims to fill this gap by introducing 
\section{Methodology}
\label{sec:method}
\begin{figure*}
\centering
\includegraphics[width=\textwidth]{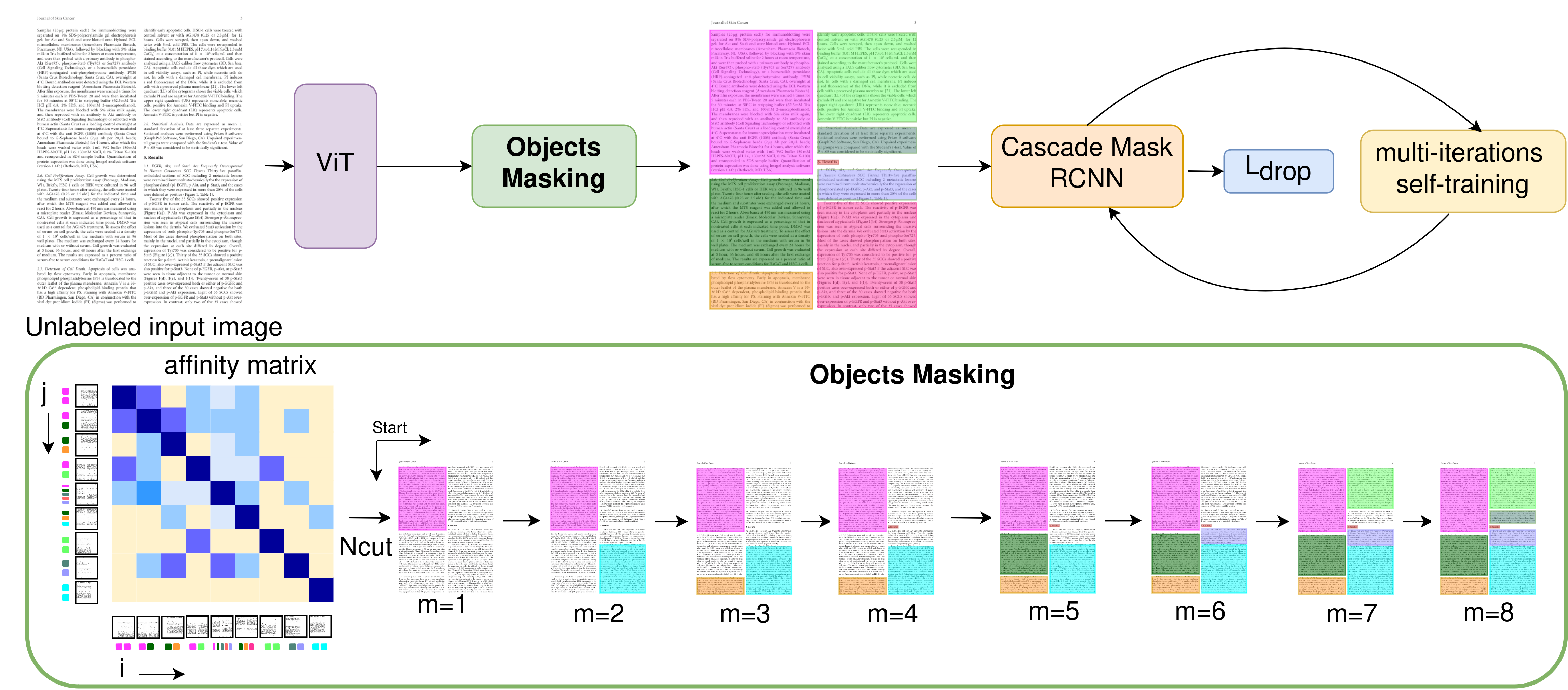}
\caption{Overview of our unsupervised training module: It takes unlabeled data to train models for object detection and instance segmentation. Then, Objects Masking~\cite{tokencut_TPAMI23} generates rough object masks utilizing the features of self-supervised DINO~\cite{DINO_selfsup3}. We employ a patch-wise similarity matrix for multiple object masks in an unlabeled image. Applying Normalized Cuts (Ncut) to this matrix, we initially extract a mask for a single foreground object. This procedure is repeated, altering the affinity matrix each time, allowing Objects Masking to discover multiple object masks in one image, demonstrated here with eight iterations.}\label{fig:cutler_method}
\end{figure*}
In our research, we focus on applying unsupervised learning to document layout segmentation and object detection domains, as shown in Fig.~\ref{fig:cutler_method}. Our primary data, denoted as \(\mathcal{D}\), consists of a comprehensive collection of RGB document images. To align with the unsupervised learning framework, which emphasizes learning from unlabeled data, we derive an unlabeled dataset \(\mathcal{D}_u = \{x_u^i\}_{i=1}^{N_u}\) from \(\mathcal{D}\), where \(N_u\) represents the total number of images in \(\mathcal{D}_u\). It does not contain traditional annotations or labels usually associated with supervised learning tasks, such as explicit object categories, locations, or dimensions. 

Initially, we employ a mask generation technique following~\cite{cutler_CVPR23,tokencut_TPAMI23,DINO_selfsup3} that creates several binary masks for each document image utilizing unsupervised features derived from DINO~\cite{DINO_selfsup3}. The approach for extracting this mask is detailed in Section~\ref{sec:mask}, highlighting the extraction process that emphasizes the document's physical layout. Furthermore, as outlined in Section~\ref{sec:loss},  we employ a dynamic loss reduction approach to effectively train a detector using the initial masks generated previously while simultaneously prompting the model to identify object masks that may have been overlooked. Lastly, as explained in Section~\ref{sec:itr}, we enhance our method's effectiveness by implementing several iterations of unsupervised training.
\subsection{Layout Mask Generation for Multiple Objects}
\label{sec:mask}
Generating the layout masks is crucial in our approach, as our unsupervised framework relies on them for visual guidance. For input document image x, we create multiple object masks within an image without the need for any manual annotations. In our approach, we initially partition the input document image into smaller image patches. We create a patch-wise similarity matrix to analyze the relationships between these patches. The crucial aspect here is using a self-supervised DINO~\cite{DINO_selfsup3}, which extracts meaningful features from these patches without needing labeled data. These extracted features are then employed to determine the similarity between each pair of patches, resulting in the formation of the similarity matrix as follows: 

\begin{equation}
  W_{ij} = \frac{F_i F_j}{\| F_i \|^2 \| F_j \|^2} 
  \label{eq:sim}
\end{equation}
where $F_i$ and $F_j$ represent the key features of patch i and patch j, respectively. The diagonal elements in the patch-wise similarity matrix have the highest values because they represent the same patch overlapping with itself, making them inherently identical and, therefore, maximally similar as shown by arrows around infinity metrix in Fig~\ref{fig:cutler_method}. This matrix is a fundamental component in our pipeline, facilitating subsequent analysis and tasks by capturing the visual relationships within the document image. We then employ the Normalized Cuts algorithm~\cite{Ncuts_TPAMI2000} on the similarity matrix, generating a single mask that highlights the primary foreground object within the image. Normalized Cuts (NCut) approaches consider image segmentation a problem of dividing a graph into meaningful parts. To do this, we create a fully interconnected and undirected graph, representing each image patch as a node. Edges between nodes are established with weights, denoted as $W_{ij}$, which quantify how similar the connected nodes are. NCut aims to find the optimal way to split this graph into two distinct sub-graphs, essentially forming a bipartition. It is achieved by solving a generalized eigenvalue system, minimizing the overall cost of this partitioning process as follows:

\begin{equation}
 (D^m - W)x^m = \lambda D^mx^m 
 \label{eq:D}
\end{equation}
where \( x^m \) is the eigenvector associated with the second smallest eigenvalue \( \lambda \) at stage m. Here, \( D^m \) represents a diagonal matrix of size \( N \times N \), with \( d(i) = \sum_j W_{ij} \), and \( W \) is a symmetrical matrix of size \( N \times N \).  One crucial aspect of this approach is determining which group of patches corresponds to the foreground, a fundamental step in object mask generation. For this, we employ two specific criteria. Firstly, we identify the patch with the highest absolute value in the second smallest eigenvector of the binary mask $M^m$. This selection intuitively represents the most prominent part of the foreground, enhancing object detection. Secondly, we incorporate a straightforward yet empirically effective prior: the foreground group should not contain two of the four input image corners. These criteria help ensure accurate identification of the foreground and background regions. The generated mask for a single document object is as follows:
\begin{equation}
    M^m_{ij} = 
    \begin{cases}
        1, & \text{if } M^m_{ij} \geq \text{mean}(x^m) \\
        0, & \text{otherwise}.
    \end{cases}
\end{equation}
where, If $M^m_{ij}$ is greater than or equal to the average value of \( x^m \), it sets $M^m_{ij}$ to 1, effectively marking that element in the mask. If $M^m_{ij}$ is less than the mean of $x^m$, it sets $M^m_{ij}$ to 0, indicating that the element is not part of the mask. In this way, it generates a mask that identifies elements belonging to the foreground.
If we don't meet certain criteria previously explained, as if there are two input image corners in the current foreground, We reverse the foreground and background as $M^m_{ij}= 1- M^m_{ij}$. Moreover, we set values of $W_{ij}$ less than $\tau_{t}$ to $1 \times 10^{-5}$ and values greater than or equal to $\tau_{t}$ to 1. 

\noindent\textbf{Mask Pooling:}
To ensure that each object in the sequence receives a distinct mask, focusing on different data or image areas. We exclude nodes previously identified as part of the foreground. This exclusion ensures that the mask generation process remains consistent with the specific characteristics of each object, leading to accurate mask generation. For this, we obtain the mask for the \((m+1)_{th}\) object by updating the node similarity \(W^{m+1}_{ij}\) and excluding the nodes corresponding to the foreground in previous stages as follows:
\begin{equation}
W^{m+1}_{ij} = \frac{(F_i \prod_{l=1}^{m} \hat{M}^l_{ij})(F_j \prod_{l=1}^{m} \hat{M}^l_{ij})}{\|F_i\|_2 \|F_j\|_2}
\end{equation}
were,\(\hat{M}^l_{ij} 1- {M}^l_{ij}\). Here, masking by excluding the nodes of previously masked foreground enables our approach to uncover multiple object masks within a single image. In document mask generation, we've set m to 10. We can vary this according to maximum possible objects in the document image. In Fig~\ref{fig:cutler_method} we adept at generating up to six distinct object masks in the image. This strategic masking enables the uncovering of multiple object masks within a single image. Employing the updated similarity matrix $W_{m+1,ij}$, we iterate through Eqs.~\ref{eq:sim} and~\ref{eq:D} to derive a new mask denoted as $M^{m+1}$. This innovative pipeline allows us to reveal and distinguish various objects within the same image without manual supervision or annotations.

\noindent\textbf{Augmentation:}
In our training process, we incorporate copy-paste augmentation approach, following~\cite {copy_paste45,dwibedi2017cut}. However, we modify this technique to enhance our model's ability to segment small objects precisely. Traditionally, copy-paste augmentation involves taking a portion of an image and placing it elsewhere within the same image or in another image. Instead of following this conventional approach, we introduce an additional step. When we copy a portion of the mask, we randomly reduce its size by a certain factor. This reduction is determined by a scalar value that we randomly select from a uniform distribution between 0.3 and 1.0. For small objects, we downsizing the mask this way to effectively replicate scenarios where objects are small. This adjustment aids the model in becoming more proficient at handling and accurately segmenting these smaller objects throughout its training process, leading to an overall enhancement in its performance.
\subsection{Loss Reduction for Exploring Object Regions}
\label{sec:loss}
 In standard object detection, the loss function penalizes predictions \( p_j \) that do not align with the actual ground-truth. However, in our unsupervised setting, we consider the previously generated mask as the ground-truth that may overlook certain instances, making it essential to extend beyond the standard loss to enable the detector to identify new, unlabeled instances effectively. To address this challenge, we employ $L_{\text{drop}}$, which selectively ignores the loss for predicted regions ($p_j$) that exhibit minimal overlap with the masked ground-truth. During training, we drop the loss for each predicted region ($p_j$) if its maximum Intersection over Union (IoU) with any masked ground-truth instance is below a threshold of $\tau_{i}=0.01$, as described by the equation:
\begin{equation}
L_{\text{drop}}(p_j) = 
\begin{cases}
L_{\text{det}}(p_j) & \text{if } \text{IoU}_j^{\text{max}} > \tau_{i}=0.01 \\
0 & \text{otherwise}
\end{cases}
\end{equation}
Here, $\text{IoU}_j^{\text{max}}$ represents the highest IoU of $p_j$ with all generated masked instances, and $L_{\text{det}}$ denotes the conventional loss function used in detectors. By implementing $L_{\text{drop}}$, the model avoids penalties for detecting objects not present in the previously generated mask, allowing it to focus on exploring various image regions.

\subsection{Multi-Iterations Unsupervised Training}
\label{sec:itr}
Our experiments show that as we train detection models, they become surprisingly good at improving the quality of the masks they generate. Even when they start with rough masks, the models gradually make them better. It, along with $L_{\text{drop}}$ strategy, helps the models find new object masks effectively. To improve performance, we employ multiple rounds of unsupervised training. We take the masks and proposals generated in the previous round in each round, but only if they have a confidence score exceeding \( 0.75 - 0.5 \) from the \( m \)-th round. These become annotations for the next round \((m + 1)\)-th, helping the model learn more about the objects in the data. To avoid feeding the network redundant information, we skip ground-truth masks that have IoU greater than 0.5 with the predicted masks. We aim to avoid redundancy in the model's learning process to ensure efficiency. Our experiments have shown that doing this training process three times works well. With each round, the model has more high-quality mask examples to learn from, making it better at generating object masks in complex scenes.

%\begin{figure*}
%\centering
%\includegraphics[width=0.9\textwidth]{images/cutler_vis.png}
%\caption{Visual analysis of our unsupervised learning approach using Cascade Mask RCNN on the %PubLayNet Dataset.}\label{fig:encoder-decoder}
%\end{figure*}

\section{Experimental Setup}
\label{sec:exp}
\subsection{Datasets}
We employ several specialized datasets such as PubLayNet~\cite{PubLayNet3}, DocLayNet~\cite{doclaynet_data}, and TableBank~\cite{tablebank8} for our document unsupervised detection and segmentation framework. DocLayNet~\cite{doclaynet_data} dataset includes 69,375 training images, 6,489 validation images, and 4,999 test images across six domains, each annotated for 11 classes. PubLayNet~\cite{PubLayNet3}, a large public dataset, contains 335,703 training, 11,240 validation, and 11,405 test images, with annotations for figures, lists, titles, tables, and texts in academic images. TableBank~\cite{tablebank8} dataset is designed to identify tables in scientific documents and contains 417,000 document images from the arXiv database. It classifies tables into LaTeX, Word, and combined categories and includes table structure recognition data. However, we only used the training images without ground-truth labels during the training.

\subsection{Evaluation Metrics}
We evaluate our unsupervised document analysis approach using the following metrics: \( mAP^{box} \), \( AP^{box}_{50} \), \( AP^{box}_{75} \), \( mAP^{mask} \), \( AP^{mask}_{50} \), and \( AP^{mask}_{75} \). The mean Average Precision \( mAP^{box} \) calculates the average precision of bounding box detections. \( AP^{box}_{50} \) and \( AP^{box}_{75} \) extend this evaluation to specific IoU thresholds of 50\% and 75\%, respectively. Similarly, \( mAP^{mask} \) measures the precision of object segmentation masks, while \( AP^{mask}_{50} \) and \( AP^{mask}_{75} \) assess this precision at the same IoU thresholds. These metrics provide a comprehensive assessment of the model's capability in accurately detecting and segmenting objects with varying degrees of precision.

\subsection{Implementation Details}
Our approach employs Document analysis dataset, without utilizing any annotations during training. For image processing, Objects Masking is employed in three stages. Images are resized to \(480\times480\) pixels, and a patch-wise similarity matrix is generated using the ViT-B/8 DINO model. Post-processing of masks is conducted using a Conditional Random Field (CRF) to calculate their bounding boxes. We employ Cascade Mask R-CNN~\cite{cascadercnn8} starting with initial masks and bounding boxes for \(150k\) iterations. Specifically, when leveraging a ResNet-50 backbone~\cite{resnet45}, the model is initially equipped with weights from a self-supervised pretrained DINO model~\cite{DINO_selfsup3}. We train our network on 2 GPUs RTXA6000 for around 8 hours. The detector is optimized over \(150k\) iterations using Stochastic Gradient Descent (SGD). It begins with a learning rate of \(0.005\), which is decreased by \(5\) times after \(80k\) iterations. The training uses batches of \(16\), a weight decay of \(5\times10^{-5}\), and a momentum of \(0.9\). 
\subsection{Performance Analysis}
The effectiveness of our unsupervised training method is evaluated in Table~\ref{tab:all_data_results}. It shows unsupervised performance for object detection and instance segmentation on different datasets, PubLayNet, DocLayNet, and TableBank.
TableBank outperforms PubLayNet and DocLayNet due to its single-class focus on tables, making the task simpler. Consequently, TableBank achieves significantly higher accuracy in both bounding box and mask predictions. We initialize the backbone with DINO network~\cite{DINO_selfsup3} and employ cascade Mask RCNN as the detector. TableBank shows high AP and mAP scores, indicating precise detection and segmentation capabilities without table labels. 
\begin{figure}
\centering
\includegraphics[width=0.87\textwidth]{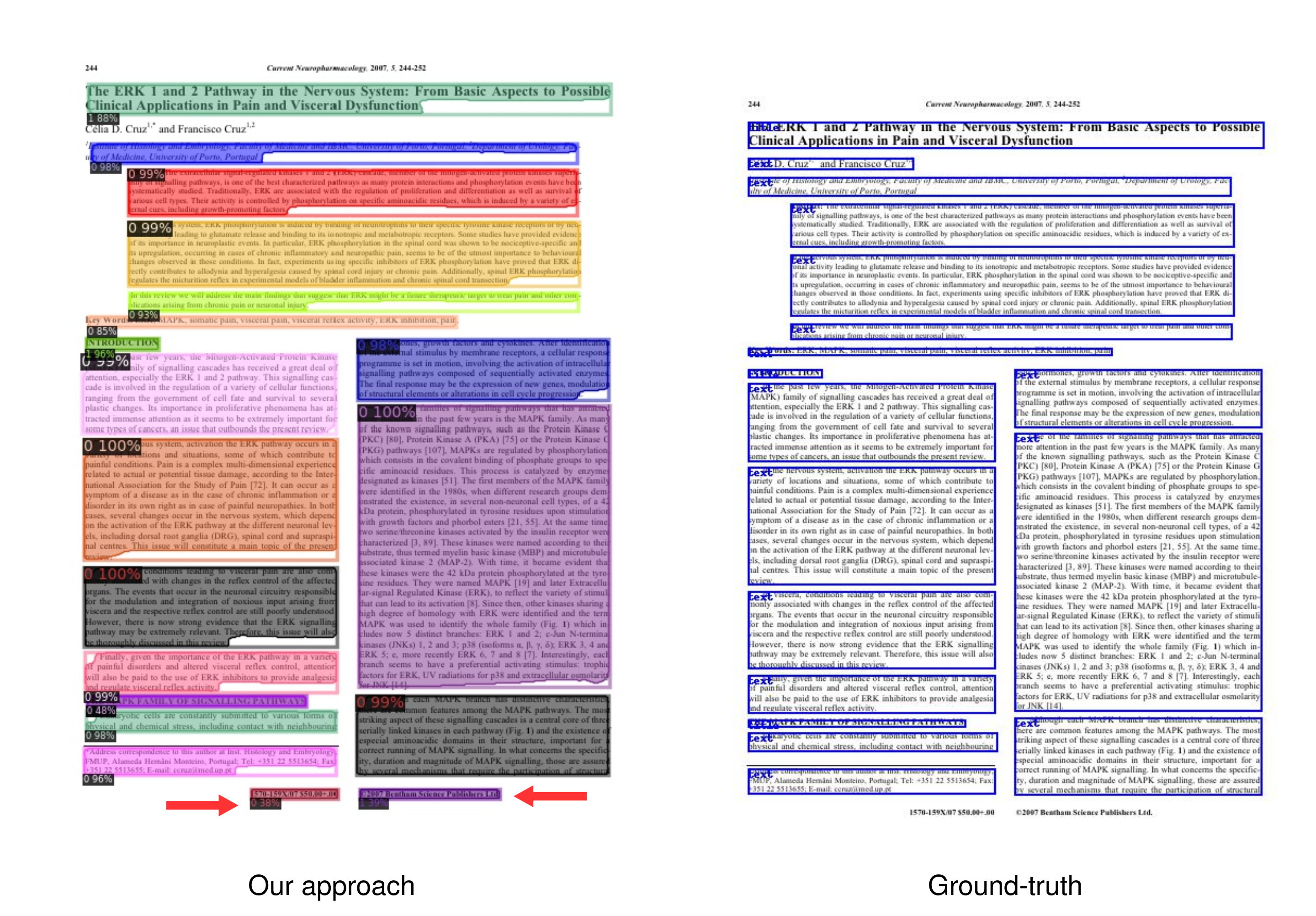}
\caption{Comparative visual analysis of unsupervised learning on the PubLayNet dataset: top-predicted layouts; bottom-corresponding ground-truth layouts. The model's proficiency in detecting details overlooked by human annotators is also highlighted, marked by red arrows.}\label{fig:visual_analysis}
\end{figure}

\begin{table}[h!]
  \centering
  \caption{Quantitative analysis of unsupervised detection and segmentation in document datasets such as PubLayNet, DocLayNet, and TableBank. We discuss the effectiveness of detection and segmentation, focusing on the detection method and backbone initialization (Init) with DINO~\cite{DINO_selfsup3}. The term 'Cascade' here represents the Cascade Mask R-CNN network~\cite{cascadercnn8}.}\label{tab:all_data_results}
  \begin{tabular}{lccccccccccc}
    \toprule
    \multirow{2}{*}{Dataset} & \multirow{2}{*}{Unsup-train} & \multirow{2}{*}{Detector} & \multirow{2}{*}{Init.} & \multicolumn{6}{c}{Performance} \\
    \cmidrule(lr){5-10}
     &  &  &  & $mAP^{box}$ & $AP^{box}_{50}$ & $AP^{box}_{75}$ & $mAP^{mask}$ & $AP^{mask}_{50}$ & $AP^{mask}_{75}$  \\
    \midrule
    PubLayNet & \multirow{3}{*}{\checkmark} & \multirow{3}{*}{'Cascade'} & \multirow{3}{*}{DINO} & 28.7 & 43.1 & 30.0 & 29.3 & 44.1 & 30.5\\
    \cmidrule(lr){5-10}
    DocLayNet &  &  &  & 22.4 & 37.5 & 23.1 & 24.2 & 38.7 & 24.8   \\
    \cmidrule(lr){5-10}
    TableBank & &  & & 88.6 & 91.2 & 89.7 & 88.8 & 91.2 & 89.7 \\
    % Add more rows as necessary
    \bottomrule
  \end{tabular}
  \label{tab:your_label}
\end{table}

TableBank has mAP of 88.6\% for detection and 88.8\% for segmentation on unsupervised training. Fig.~\ref{fig:visual_analysis} shows the performance of our unsupervised learning approach on the PubLayNet dataset. The analysis includes the unsupervised model's predicted layouts against the ground-truth layouts. Notably, the model demonstrates an improved ability to recognize various elements within a document, such as footers. It also excels in precisely segmenting smaller components like text blocks. A key aspect of this analysis is the model's remarkable performance in identifying fine details within the layouts, some of which might even be missed by human annotators. These instances, where the model's predictions positively diverge from human annotations, are specifically highlighted with red arrows. It highlights the model's advanced capability in document object detection and segmentation in unsupervised settings. 

\begin{table}[h]
\centering
\caption{Merged Results for PubLayNet, TableBank, and DocLayNet}
\label{tab:merged-results-all}
\begin{tabular}{lcccccc}
\toprule
 & \multicolumn{2}{c}{PubLayNet} & \multicolumn{2}{c}{TableBank} & \multicolumn{2}{c}{DocLayNet} \\
\cmidrule{2-3} \cmidrule{4-5} \cmidrule{6-7}
Methods & $mAP^{\text{box}}$ & $mAP^{\text{mask}}$ & $mAP^{\text{box}}$ & $mAP^{\text{mask}}$ & $mAP^{\text{box}}$ & $mAP^{\text{mask}}$ \\
\midrule
\textbf{Fully-supervised methods:} & & & & & & \\
\textcolor{gray}{V+BERT-12L}~\cite{hybridvbert} & 96.5 & - & & - & 81.0 & - \\
\textcolor{gray}{VGT}~\cite{Da_2023_ICCV} & 96.2 & - & & - & - & - \\
\textcolor{gray}{SwinDocSegmenter}~\cite{SwinDoc} & - & 93.72 & & - & 98.04 & - \\
\textcolor{gray}{TRDLU}~\cite{9897491} & 95.95 & - & & - & - & - \\
\textcolor{gray}{VSR}~\cite{VSR} & 95.7 & - & & - & - & - \\
\textcolor{gray}{CDeC-Net}~\cite{cdec} & 96.7 & - & 89.8 & - & - & - \\
\textcolor{gray}{DocSegTr}~\cite{docsegtr} & - & - & & 93.3 & - & - \\
\textcolor{gray}{Layout LMv3}~\cite{huang2022layoutlmv3} & - & - & & 92.9 & - & - \\
\textcolor{gray}{GLAM + YOLOv5x6}~\cite{glamyolo} & - & - & & - & - & 80.8 \\
\textcolor{gray}{Mask R-CNN}~\cite{Pfitzmann_2022} & - & - & & - & - & 78.0 \\
\midrule
\textbf{Unsupervised methods:} & & & & & & \\
Our & 28.7 & 29.3 & 88.6 & 88.8 & 22.4 & 24.2 \\
\bottomrule
\end{tabular}
\end{table}
Table~\ref{tab:merged-results-all} compares our unsupervised approach with previous fully supervised approaches, highlighting the effectiveness of the unsupervised approach in object detection and segmentation tasks within document layout analysis. Supervised methods, which have the advantage of learning from labeled data, generally yield high precision scores; for instance, SwinDocSegmenter~\cite{SwinDoc} achieves an impressive 93.72 $AP_{box}$ on TableBank, indicating its strong capability to identify and localize objects accurately. However, the unsupervised method is particularly noteworthy, achieving an $AP_{mask}$ of 88.8 on TableBank without the aid of labeled training data. This high score in segmentation precision suggests that our approach can predict the shapes and boundaries of document elements, such as tables or text blocks, almost as effectively as its supervised approaches. The ability of our approach to perform so well in an unsupervised manner is significant as it implies a considerable reduction in the dependency on costly and time-consuming data labeling processes. It also opens up new possibilities for analyzing documents in domains where obtaining labeled data is difficult, thus expanding the applicability of unsupervised learning in document analysis. Therefore, it provides a performance benchmark for current methods and the possibility of unsupervised learning approaches in real-world document layout understanding tasks.
\begin{table}[ht]
\centering
\begin{subtable}{.50\linewidth}
\centering
\begin{tabular}{ccccc}
\toprule
Size $\rightarrow$ & 240 & 360 & 480 & 640 \\
\midrule
$AP^{mask}_{50}$ & 86.4 & 87.5 & 88.6 & 88.7 \\
\bottomrule
\end{tabular}
\caption{Image size.}
\end{subtable}\hfill
\begin{subtable}{.50\linewidth}
\centering
\begin{tabular}{ccccc}
\toprule
$\tau_{t} \rightarrow $ & 0 & 0.1 & 0.15 & 0.2 \\
\midrule
$AP^{mask}_{50}$ & 88.2 & 88.5 & 88.6 & 88.5 \\
\bottomrule
\end{tabular}
\caption{$\tau_{\text{t}}$ for Objects Masking.}
\end{subtable}\hfill
\begin{subtable}{.5\linewidth}
\centering
\begin{tabular}{cccc}
\toprule
$N \rightarrow$ & 5 & 10 & 15 \\
\midrule
$AP^{mask}_{50}$ & 88.1 & 88.6 & 88.6 \\
\bottomrule
\end{tabular}
\caption{\# masks per image.}
\end{subtable}\hfill
\begin{subtable}{.5\linewidth}
\centering
\begin{tabular}{ccccc}
\toprule
$\tau_{i} \rightarrow$ & 0 & 0.01 & 0.1 & 0.2 \\
\midrule
$AP^{mask}_{50}$ & 88.3 & 88.6 & 85.5 & 82.9 \\
\bottomrule
\end{tabular}
\caption{$\tau_{i}$ for $L_{\text{drop}}$.}
\end{subtable}\vspace{1pt}
\caption{Ablations for mask generation and loss reduction for exploring object regions. This study examines the impact of different parameters on unsupervised training performance using the TableBank dataset. The parameters varied include: (a) image size, (b) the threshold value $\tau_{t}$ which determines the sparsity level of the affinity matrix in Normalized Cuts, (c) the number of masks generated by Objects Masking, and (d) the threshold $\tau_{i}$ in $L_{\text{drop}}$, which is the maximum allowable overlap between predicted regions and ground-truth before excluding loss for those regions. Default parameter settings are indicated in gray.}\label{tab:parameters}
\end{table}
\section{Ablation Study}
\label{sec:AS}
\noindent\textbf{Design choices of unsupervised training parameters.}
This study conducts an ablation analysis on the design choices of unsupervised training parameters in the context of mask generation and loss reduction for exploring object regions, as shown in Table~\ref{tab:parameters}. The research is centered on utilizing the TableBank dataset to evaluate the impact of various parameters on unsupervised training performance. The parameters under scrutiny encompass: (a) the image size, (b) the threshold value $\tau_{t}$, which plays a crucial role in determining the sparsity of the affinity matrix within the Normalized Cuts method, (c) the quantity of masks generated through the Objects Masking technique, and (d) the threshold $\tau_{i}$ in $L_{\text{drop}}$, which dictates the maximum allowable overlap between predicted regions and ground-truth before dismissing the loss for those regions. A key aspect of this analysis is identifying default parameter settings, which are distinctly highlighted in gray for reference. Understanding the influence of unsupervised training parameters in object region exploration is important for optimizing mask generation and loss reduction efficiency and accuracy. By varying these parameters and assessing their effects on performance, this research provides the best results for enhancing the overall performance. The study's insights can aid in fine-tuning unsupervised training processes, ensuring more precise and effective results in tasks related to document analysis and object recognition. 

\noindent\textbf{Effectiveness of unsupervised training iterations.}
Multiple rounds of unsupervised training effectively enhance the quality and quantity of object masks, as indicated in Table~\ref{tab:self_training_rounds}. Through iterative refinement, the model progressively improves the precision of object masks, even when starting with rough initial predictions. This process generates more masks, aiding the model's training. Combining these masks with the $L_{\text{drop}}$ strategy, which focuses on uncertain predictions, helps the model target areas where it initially struggles, improving mask accuracy. Our experiments suggest that performing unsupervised training three times provides a balance between generating high-quality masks and avoiding overfitting, making it particularly valuable for handling even small and complex document objects in document analysis data.

\begin{table*}[htp!]
\begin{center}
\begin{minipage}[b]{.70\textwidth}
\caption{Analysis of training iterations in unsupervised learning. Here, analysis shows that three iterations provide the best results using Cascade Mask RCNN on the TableBank dataset.}
\label{tab:self_training_rounds}
\renewcommand{\arraystretch}{1}
\begin{tabular*}{\textwidth}
{@{\extracolsep{\fill}}cclllll@{\extracolsep{\fill}}}
\toprule
Iteration  & \( mAP^{box} \) & \( AP^{box}_{50} \) & \( AP^{box}_{75} \)& \( mAP^{mask} \) & \( AP^{mask}_{50} \) &  \( AP^{mask}_{75} \)\\
\midrule
1   & 86.2  & 89.5 & 88.4 & 88.2 & 89.6 & 88.9 \\
2   &  88.3 & 90.7 & 89.1 & 88.5 & 90.8 & 89.4 \\ 
3   &  88.6 & 91.2 & 89.7 & 88.8 & 91.2 & 89.7 \\ 
4   & 88.6 & 91.0 & 89.5 & 88.7 & 91.2 & 89.7 \\
\bottomrule
\end{tabular*}
\end{minipage}
\end{center}
\end{table*} 
\vspace{-10pt}
\noindent\textbf{Effectiveness of quantity of pre-training data.}
The quantity of unsupervised training data significantly influences the effectiveness of our unsupervised approach.  Essentially, the larger the dataset we have for training, the better our model tends to perform in terms of its ability to generalize and achieve higher performance. This relationship between data quantity and model performance is demonstrated in Table~\ref{tab:dataset_impact}. Using only 10\% of the data for unsupervised training, we achieved an mAP of 82.9 for detection and 85.2 for segmentation. However, when we utilized the full 100\% of the available data, our performance improved significantly to an mAP of 88.6 for detection and 88.8 for segmentation.

\begin{table*}[htp!]
\begin{center}
\begin{minipage}[b]{.70\textwidth}
\caption{Performance analysis of Cascade Mask RCNN unsupervised training with varying percentages of data utilized in TableBank dataset.}
\label{tab:dataset_impact}
\renewcommand{\arraystretch}{1}
\begin{tabular*}{\textwidth}
{@{\extracolsep{\fill}}cclllll@{\extracolsep{\fill}}}
\toprule
\% data & \( mAP^{box} \) & \( AP^{box}_{50} \) & \( AP^{box}_{75} \) & \( mAP^{mask} \) & \( AP^{mask}_{50} \) & \( AP^{mask}_{75} \) \\
\midrule
10\% & 82.9 & 88.9 & 86.0 & 85.2 & 88.9 & 86.6\\
30\% & 85.4 & 89.3 & 87.2 & 86.2 & 89.3 & 87.2 \\
50\% & 85.8 & 90.5 & 88.1 & 87.3 & 90.5 & 88.2 \\
100\%&  88.6 & 91.2 & 89.7 & 88.8 & 91.2 & 89.7  \\
\bottomrule
\end{tabular*}
\end{minipage}
\end{center}
\end{table*} 
\noindent\textbf{Effectiveness of cross-data unsupervised learning.}
Moreover, in Table~\ref{tab:cross-data}, we examine the impact of training data on the efficiency of unsupervised training. Specifically, we investigate the performance differences when a network is unsupervisely trained sequentially on two distinct datasets. Initially, the network undergoes unsupervised training for 150k iterations exclusively on just PubLayNet dataset. In second experiment, the network is first unsupervisely trained on the TableBank dataset for 75k iterations.
\begin{table*}[htp!]
\begin{center}
\begin{minipage}[b]{.95\textwidth}
\caption{Impact of Dataset Selection on Cross Unsupervised Training. We explore how different datasets affect cross unsupervised training results.}\label{tab:cross-data}
\renewcommand{\arraystretch}{1} % Default value: 1
\begin{tabular*}{\textwidth}
{@{\extracolsep{\fill}}cclllll@{\extracolsep{\fill}}}
\toprule
Cross Unsup-training & \( mAP^{box} \) & \( AP^{box}_{50} \) & \( AP^{box}_{75} \)& \( mAP^{mask} \) & \( AP^{mask}_{50} \) & \( AP^{mask}_{75} \)\\
\midrule
 PubLayNet & 28.7 & 43.1 & 30.0 & 29.3 & 44.1 & 30.5 \\
TableBank $+$ PubLayNet & 65.6 & 84.8 & 71.2 & 65.3 & 85.2 & 71.5 \\
\bottomrule
\end{tabular*}
\end{minipage}
\end{center}
\end{table*} 
Following this, the network undergoes an additional 75k iterations of unsupervised training on the PubLayNet dataset. Our findings reveal a significant performance improvement when cross-training is employed. Specifically, training solely on the PubLayNet dataset resulted in a mAP of 28.7 for document object detection. In contrast, the cross-data training approach, involving both TableBank and PubLayNet datasets, yields a substantially higher mAP of 65.6. Our experiments show that unsupervised training the network on multiple datasets, rather than just one, significantly improves its performance.
\section{Conclusion}
\label{sec:conclusion}
In conclusion, the paper presents a significant advancement in the field of document layout analysis by introducing a vision-based approach that effectively addresses the challenges of limited labeled data and the diversity of documents online. This method diverges from traditional techniques that rely heavily on labeled data, which are increasingly impractical due to the massive volume of documents on the internet. The proposed approach begins with pre-training that generates simple object masks from unlabeled document images, bypassing the need for extensive labeling. These masks are then employed to train a detector, leading to improved object detection and segmentation precision. The model's performance is further enhanced through multiple training iterations, allowing for continuous refinement. This approach offers a more efficient, accurate, and flexible way for analyzing document layouts, making a major improvement in the field of document research. In the future research, we intend to investigate how unsupervised techniques can be utilized to improve Document Layout Analysis.
\section*{Acknowledgements}
The work leading to this publication has been partially funded by the EU Horizon Europe Project AIRISE (https://airise.eu/) under grant agreement 101092312.

\bibliographystyle{IEEEtran}
\bibliography{main}% common bib file

\end{document}